%% file: main.tex
\PassOptionsToPackage{table}{xcolor}
\documentclass[10pt,twocolumn,letterpaper]{article}

\usepackage{iccv}
\usepackage{url}
\usepackage[bb=boondox]{mathalfa}
\usepackage{graphicx}
\usepackage{amsmath,amsthm}
\usepackage{amssymb}
\usepackage{booktabs}
\usepackage{multirow}
\usepackage{times}
\usepackage{tikz}
\usepackage{comment}
\usepackage{wrapfig}
\usepackage{lipsum}
\input{math_commands.tex}

\usepackage[normalem]{ulem}

\usepackage[pagebackref=true,breaklinks=true,letterpaper=true,colorlinks,bookmarks=false,pdfstartview={FitH}]{hyperref}

\iccvfinalcopy 

\newcommand{\improvecolor}{\color[HTML]{3B9612}}

\def\iccvPaperID{3923} 

\ificcvfinal\fi

\begin{document}

\title{Reducing Training Time in Cross-Silo Federated Learning using \\ Multigraph Topology}
\author{Tuong Do$^{1\dagger}$, Binh X. Nguyen$^{1\dagger}$, Vuong Pham$^{1}$, Toan Tran$^{2}$,\\Erman Tjiputra$^{1}$, Quang D. Tran$^{1}$, Anh Nguyen$^{3}$\\
{$^{1}$AIOZ, Singapore}\\
{$^{2}$VinAI Research, Vietnam}\\
{$^{3}$University of Liverpool, UK}\\
{\tt\small \{tuong.khanh-long.do, binh.xuan.nguyen\}}
{\tt\small @aioz.io}\\
}

\maketitle
\begin{abstract}
Federated learning is an active research topic since it enables several participants to jointly train a model without sharing local data. Currently, cross-silo federated learning is a popular training setting that utilizes a few hundred reliable data silos with high-speed access links to training a model. While this approach has been widely applied in real-world scenarios, designing a robust topology to reduce the training time remains an open problem. In this paper, we present a new multigraph topology for cross-silo federated learning. We first construct the multigraph using the overlay graph. We then parse this multigraph into different simple graphs with isolated nodes. The existence of isolated nodes allows us to perform model aggregation without waiting for other nodes, hence effectively reducing the training time. Intensive experiments on three public datasets show that our proposed method significantly reduces the training time compared with recent state-of-the-art topologies while maintaining the accuracy of the learned model. Our code can be found at:\url{https://github.com/aioz-ai/MultigraphFL}
\end{abstract}

\section{Introduction}
\let\thefootnote\relax\footnotetext{$\dagger$ \textrm{\ indicates\ equal\ contribution.}}
\begin{figure}[ht]
  \begin{center}
    \includegraphics[width=0.9\textwidth, height=0.7\textwidth]{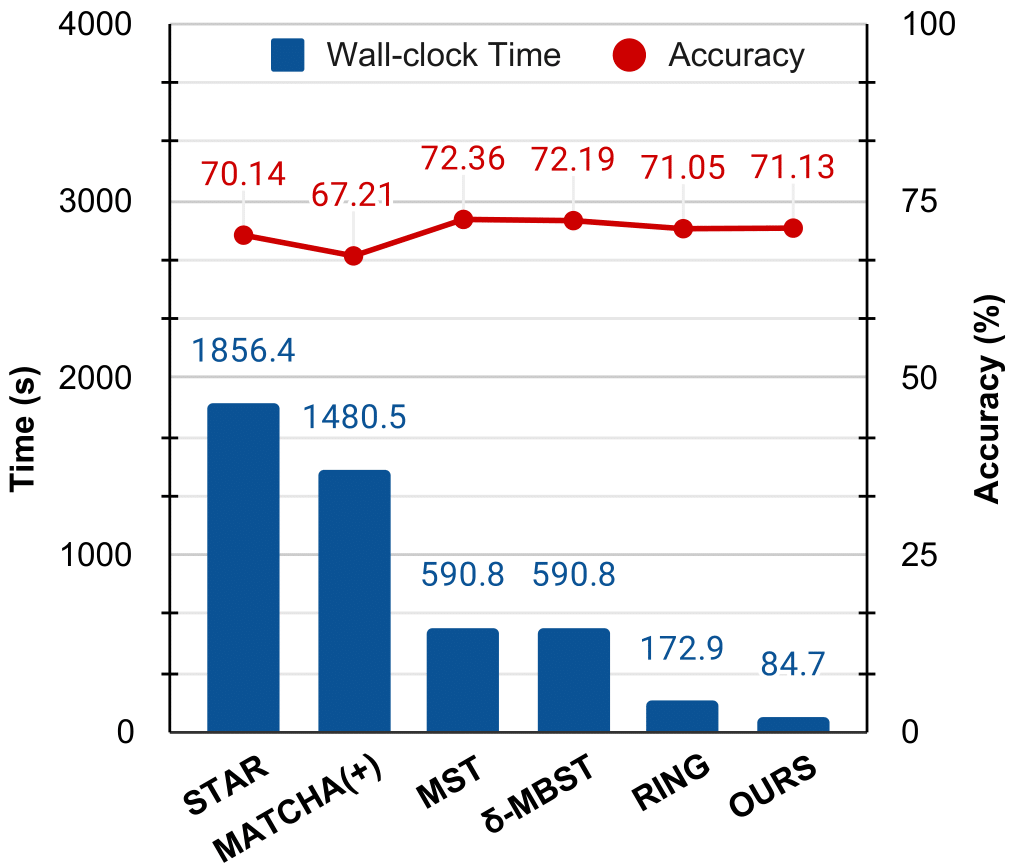}
  \end{center}
  \caption{Comparison between different topologies on FEMNIST dataset and Exodus network~\cite{awscloud}. The accuracy and total wall-clock training time (or overhead time) are reported after $6,400$ communication rounds. 
  Our method significantly reduces the training time while maintaining the model accuracy.}
  \vspace{-15pt}
  \label{fig:Intro_Performance}
\end{figure}
\vspace{-1ex}

Federated learning entails training models via remote devices or siloed data centers while keeping data locally to respect the user's privacy policy~\cite{li2020federated}. According to~\cite{kairouz2019advances}, there are two popular training scenarios: the \textit{cross-device} scenario, which encompasses a variety (millions or even billions) of unreliable edge devices with limited computational capacity and slow connection speeds; and the \textit{cross-silo} scenario, which involves only a few hundred reliable data silos with powerful computing resources and high-speed access links. Recently, the cross-silo scenario becomes popular in different federated learning applications such as
healthcare~\cite{xu2021federated,joshi2022federated,hosseini2023proportionally}, robotics~\cite{nguyen2021deep,zhang2021distributed,wang2022atpfl}, medical imaging~\cite{courtiol2019deep,liu2021feddg}, and finance~\cite{shingi2020federated,liu2023federated}. 

In practice, federated learning is a promising research direction where we can utilize the effectiveness of machine learning methods while respecting the user’s privacy. Critical challenges in federated learning include model convergence, communication congestion, and imbalance of data distributions in different silos~\cite{kairouz2019advances}. A popular federated training method is to set a central node that orchestrates the training process and aggregates contributions of all clients~\cite{mcmahan2017communication}. The main limitation of this client-server approach is that the server node potentially represents a communication congestion point in the system, especially when the number of clients is large. To overcome this limitation, recent research has investigated the decentralized (or peer-to-peer) federated learning approach~\cite{lian2018asynchronous,he2019central,marfoq2020throughput,li2021decentralized,jeong2023personalized}, in which the communication is performed via a peer-to-peer topology without the need for a central node. However, the main challenge of decentralized federated learning is to achieve fast training time while assuring model convergence and maintaining model accuracy.

In federated learning, the communication topology plays an important role. In particular, an efficient topology leads to faster convergence and reduces the training time and energy usage, quantifying by the worst-case convergence bounds in the topology design~\cite{jiang2017collaborative,nedic2018network,wang2018cooperative}. Furthermore, topology design is directly related to other problems during the training process such as network congestion, the overall accuracy of the trained model, or energy consumption~\cite{yang2021achieving,nguyen2021deep,kang2019incentive}. Designing a robust topology that can reduce the training time while maintaining the model accuracy is still an open problem in federated learning~\cite{kairouz2019advances}. Our paper aims to design a new topology for cross-silo federated learning, the most common training scenarios in practice.  

Recently, different topologies have been proposed for cross-silo federated learning. In~\cite{brandes2008variants}, the STAR topology is designed where the orchestrator averages all models throughout each communication round. The authors in~\cite{wang2019matcha} propose MATCHA to decompose the set of possible communications into pairs of clients. At each communication round, they randomly select some pairs and allow them to transmit models. The authors in~\cite{marfoq2020throughput} introduce RING topology using max-plus linear systems. While some progress has been made in the field, there are challenging problems that need to be addressed such as congestion at access links~\cite{wang2019matcha, yang2021achieving}, straggler effect~\cite{neglia2019role, park2021sageflow}, or identical topology in all communication rounds~\cite{jiang2017collaborative,marfoq2020throughput}. 

In this paper, we propose a new multigraph topology based on the recent RING topology~\cite{marfoq2020throughput} to reduce the training time for cross-silo federated learning. Our method first constructs the multigraph based on the overlay of the RING topology. We then parse this multigraph into simple graphs containing only one edge between two nodes. We refer to each simple graph as a \textit{state} of the multigraph. Each state of the multigraph may have isolated nodes, and these nodes can do model aggregation without waiting for other nodes. This strategy significantly reduces the cycle time in each communication round. 
Intensive experiments show that our proposed topology outperforms other state-of-the-art methods in terms of training time for cross-silo federated learning by a large margin (See Figure~\ref{fig:Intro_Performance}).
\section{Literature Review}
\label{Sec:Literature}
\textbf{Federated Learning}. Federated learning has been regarded as a system capable of safeguarding data privacy~\cite{konevcny2016federated,gong2021ensemble,zhang2021federated,li2021model,gao2022feddc}.
Contemporary federated learning is based on a centralized network design in which a central node receives gradients from the client nodes to update a global model. Early findings of federated learning research include the work of~\cite{konevcny2015federatedB}, as well as a widely circulated article from~\cite{mcmahan2017google}. Then ~\cite{yang2013analysis,shalev2013stochastic,ma2015adding,jaggi2014communication}, and ~\cite{smith2018cocoa} extend the concept of federated learning and its related distributed optimization algorithms. Federated Averaging (FedAvg) was proposed by~\cite{mcmahan2017communication}, its variations such as FedSage~\cite{zhang2021subgraph} and DGA~\cite{zhu2021delayed}, or other recent state-of-the-art model aggregation methods~\cite{hong2021efficient,ma2022layer,zhang2022fine,liu2022deep,elgabli2022fednew,jain2023federated,chen2023federated} are introduced to address the convergence and non-IID (non-identically and independently distributed) problem. Despite its simplicity, the client-server approach suffers from the communication and computational bottlenecks in the central node, especially when the number of clients is large~\cite{he2019central,qu2022rethinking}.

\textbf{Decentralized Federated Learning}. Decentralized (or peer-to-peer) federated learning allows each silo data to interact with its neighbors directly without a central node~\cite{he2019central}. Due to its nature, decentralized federated learning does not have communication congestion at the central node, however, optimizing a fully peer-to-peer network is a challenging task~\cite{nedic2014distributed,lian2017can,he2018cola,lian2018asynchronous,wang2019matcha,marfoq2020throughput,marfoq2021federated,li2021decentralized}. Noticeably, the decentralized periodic averaging stochastic
gradient descent~\cite{wang2018cooperative} is proved to converge at a comparable rate to the centralized algorithm while allowing large-scale model training~\cite{wu2017decentralized,shen2018towards,odeyomi2021differentially}. 
 Besides, systematic analysis of the decentralized federated learning has been explored by~\cite{li2018pipe,ghosh2020efficient,koloskova2020unified}. Recently, Jeong \etal~\cite{jeong2023personalized} leverage knowledge distillation mechanism to ensure the collaboration between silos in the decentralized federated scenario while preserving the privacy between neighbor nodes.  

\textbf{Communication Topology}. The topology has a direct impact on the complexity and convergence of federated learning~\cite{chen2020wireless}. 
  Many works have been introduced to improve the effectiveness of topology, including star-shaped topology~\cite{brandes2008variants,konevcny2016federated,mcmahan2016federated,mcmahan2017communication,kairouz2019advances} and optimized-shaped topology~\cite{neglia2019role,wang2019matcha,marfoq2020throughput,bellet2021d,vogels2021relaysum,huang2022tackling}. Particularly, a spanning tree topology based on~\cite{prim1957shortest} algorithm was introduced by~\cite{marfoq2020throughput} to reduce the training time. 
 As mentioned by~\cite{brandes2008variants}, STAR topology is designed where an orchestrator averages model updates in each communication round. The authors in~\cite{wang2019matcha} introduce MATCHA to speed up the training process through decomposition sampling. Since the duration of a communication round is dictated by stragglers effect~\cite{karakus2017straggler,li2018near}, ~\cite{neglia2019role} explore how to choose the degree of a regular topology. RING topology is proposed in~\cite{marfoq2020throughput} for cross-silo federated learning using the theory of max-plus linear systems. Huang~\etal~\cite{huang2022tackling} introduce sample-induced topology which is able to recover the effectiveness of existing SGD-based algorithms along with their corresponding rates. Recently, Wu~\etal~\cite{wu2023topology} conducts a comprehensive survey of models, frameworks, and algorithms on network topologies.

\textbf{Multigraph.} The definition of multigraph has been introduced as a traditional paradigm in math~\cite{gibbons1985algorithmic,walker1992implementing}. A typical ``graph" usually refers to a simple graph with no loops or multiple edges between two nodes. Different from a simple graph, a multigraph allows multiple edges between two nodes. In deep learning, multigraph has been applied in different domains, including clustering~\cite{martschat2013multigraph,luo2020deep,kang2020multi}, medical image processing~\cite{liu2018multi,zhao2021multi,bessadok2021brain}, traffic flow prediction~\cite{lv2020temporal,zhu2021spatiotemporal}, activity recognition~\cite{stikic2009multi}, recommendation system~\cite{tang2021dynamic}, and cross-domain adaptation~\cite{ouyang2019learning}. In this paper, we construct the multigraph to enable isolated nodes and reduce the training time in cross-silo federated learning. 

\section{Preliminaries}
\label{Sec:Method}
\subsection{Federated Learning}
In federated learning, silos do not share their local data, but still periodically transmit model updates between them. 
Given $N$ siloed data centers, the objective function for federated learning is:
\vspace{-2.5ex}
\begin{equation}
\min_{\textbf{w} \in \mathbb R^d}
\sum^{N}_{i=1}p_i \mathds{E}_{\xi_i}\left[ L_{i}\left(\textbf{w}, \xi_i\right)\right],
\label{eq:ori_FL}
\vspace{-1ex}
\end{equation}
where $L_{i}(\textbf{w}, \xi_i)$ is the loss of model parameterized by the weight $\textbf{w} \in \mathbb R^d$, $\xi_i$ is an input sample drawn from data at silo $i$, and the coefficient $p_i>0$ specifies the relative importance of each silo. Recently, different distributed algorithms have been proposed to optimize Eq.~\ref{eq:ori_FL}~\cite{konecny2016federated,mcmahan2017communication,li2018federated,wang2019matcha,li2019communication,wang2018cooperative,karimireddy2020scaffold}. In this work, DPASGD~\cite{wang2018cooperative} is used to update the weight of silo $i$ in each training round as follows:
\vspace{-1.5ex}
\begin{multline}
\textbf{w}_{i}\left(k + 1\right) = \\
\begin{cases}
    \sum_{j \in \mathcal{N}_i^{+} \cup{\{i\}}}\textbf{A}_{i,j}\textbf{w}_{j}\left(k\right), \\\qquad\qquad\qquad\qquad\qquad \text{if k} \equiv 0 \left(\text{mod }u + 1\right),\\
    \textbf{w}_{i}\left(k\right)-\alpha_{k}\frac{1}{b}\sum^b_{h=1}\nabla L_i\left(\textbf{w}_{i}\left(k\right),\xi_i^{\left(h\right)}\left(k\right)\right),  \\\qquad\qquad\qquad\qquad\qquad\qquad\qquad\quad\text{otherwise.}
\end{cases}
\label{eq:ori_DFL}
\vspace{-2ex}
\end{multline}
where $b$ is the batch size, $i,j$ denote the silo, $u$ is the number of local updates, $\alpha_k > 0$ is a potentially varying learning rate at $k$-th round, $\textbf{A} \in \mathds{R}^{N \times N}$ is a consensus matrix with non-negative weights, and $\mathcal{N}_i^{+}$ is the in-neighbors set that silo $i$ has the connection to.

\subsection{Multigraph for Federated Learning}

\textbf{Connectivity and Overlay}. Following~\cite{marfoq2020throughput}, we consider the \textit{connectivity} $\mathcal{G}_c = (\mathcal{V}, \mathcal{E}_c)$ as a graph that captures possible direct communications among silos. Based on its definition, the connectivity is often a fully connected graph and is also a directed graph. 
The \textit{overlay} $\mathcal{G}_o$ is a connected subgraph of the connectivity graph, i.e., $\mathcal{G}_o = (\mathcal{V}, \mathcal{E}_o)$, where $\mathcal E_o \subset \mathcal E_c$. Only nodes directly connected in the overlay graph $\mathcal{G}_o$ will exchange the messages during training. We refer the readers to~\cite{marfoq2020throughput} for more in-depth discussions.

\textbf{Multigraph}. While the connectivity and overlay graph can represent different topologies for federated learning, one of their drawbacks is that there is only one connection between two nodes. In our work, we construct a \textit{multigraph} $\mathcal{G}_m = (\mathcal{V}, \mathcal{E}_m)$ from the overlay $\mathcal{G}_o$. The multigraph can contain multiple edges between two nodes~\cite{chartrand2013first}. In practice, we parse this multigraph to different graph states, each state is a simple graph with only one edge between two nodes.

\begin{figure}[t]
\vspace{-1ex}
   \centering
\resizebox{1.0\textwidth}{!}{
   \subfigure[Connectivity]{\includegraphics[width=0.23\linewidth, height=0.23\linewidth]{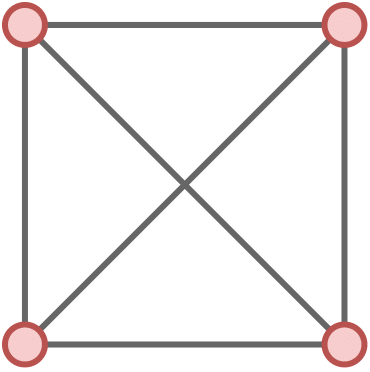}} \hspace{1ex}
   \subfigure[Overlay]{\includegraphics[width=0.23\linewidth, height=0.23\linewidth]{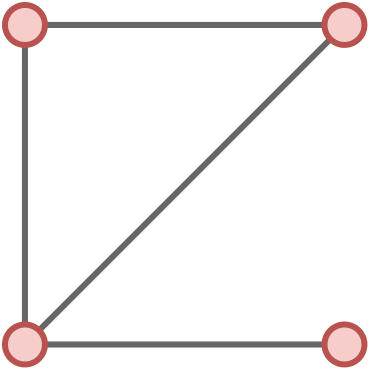}} \hspace{1ex}
   \subfigure[Multigraph]{\includegraphics[width=0.23\linewidth,, height=0.23\linewidth]{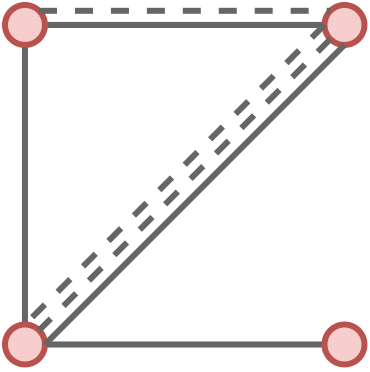}}\hspace{1ex}
   \subfigure[State of Multigraph]{\includegraphics[width=0.23\linewidth, height=0.23\linewidth]{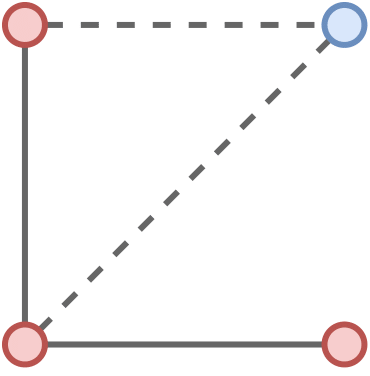}}
   }
 \caption{Example of connectivity, overlay, multigraph, and a state of our multigraph. Blue node is an isolated node. Dotted line denotes a weakly-connected edge.}
 \label{fig:SampleGraph}
\vspace{-0.5cm}
\end{figure} 

In the multigraph $\mathcal{G}_m$, the connection edge between two nodes has two types: \textit{strongly-connected} edge and \textit{weakly-connected} edge~\cite{ke2016weak}. Under both strong and weak connections, the participated nodes can transmit their trained models to their out-neighbours $\mathcal{N}_i^{-}$ or download models from their in-neighbours $\mathcal{N}_i^{+}$. However, in a strongly-connected edge, two nodes in the graph must wait until all upload and download processes between them are finished to do model aggregation. On the other hand, in a weakly-connected edge, the model aggregation process in each node can be established whenever the previous training process is finished by leveraging up-to-date models which have not been used before from the in-neighbours of that node.

\textbf{State of Multigraph}. Given a multigraph $\mathcal{G}_m$, we can parse this multigraph into different simple graphs with only one connection between two nodes (either strongly-connected or weakly-connected). We denote each simple graph as a state $\mathcal{G}_m^s$ of the multigraph. 

\textbf{Isolated Node}. A node is called isolated when all of its connections to other nodes are weakly-connected edges. Figure~\ref{fig:SampleGraph} shows the graph concepts and isolated nodes.

\subsection{Delay and Cycle Time in Multigraph}

\textbf{Delay}. Following~\cite{marfoq2020throughput}, a delay to an edge $e(i, j)$ is the time interval when node $j$ receives the weight sending by node $i$, which can be defined by:
\vspace{-1.5ex}
\begin{equation}
\small
d(i,j) = u \times T_c(i) + l(i,j) + \frac{M}{O(i,j)},
\label{eq:ori_delay} 
\vspace{-1ex}
\end{equation}
where $T_{c}(i)$ denotes the time to compute one local update of the model; $u$ is the number of local updates; $l(i,j)$ is the link latency; $M$ is the model size; $O(i, j)$ is the total network traffic capacity. However, unlike other communication infrastructures, the multigraph only contains connections between silos without other nodes such as routers or amplifiers. Thus, the total network traffic capacity $O(i,j) = \text{min}\left(\frac{C_{\rm{UP}}(i)}{\left|{\mathcal{N}_{i}^{-}}\right|}, \frac{C_{\rm{DN}}(j)}{\left|\mathcal{N}_{i}^{+}\right|}\right)$
where $C_{\rm{UP}}$ and $C_{\rm{DN}}$ denote the upload and download link capacity. Note that the upload and download processes can happen in parallel.

Since multigraph can contain multiple edges between two nodes, we extend the definition of the delay in Eq.~\ref{eq:ori_delay} to $d_k(i,j)$, with $k$ is the $k$-th communication round during the training process, as:
\vspace{-1.5ex}
\begin{multline}
d_{k+1}(i,j) =\\
\begin{cases}
    d_k(i,j), \\\qquad \qquad \text{if } e_{k+1}(i,j) = \mathbb{1}\text{ and }e_{k}(i,j) = \mathbb{1}\\
    \text{max}( u \times T_c(j),d_{k}(i,j) - d_{k-1}(i,j)), \\\qquad\qquad\text{if }e_{k+1}(i,j) = \mathbb{1}\text{ and }e_{k}(i,j) = \mathbb{0}\\
    \tau_k(\mathcal{G}_m) + d_{k-1}(i,j)), \\\qquad\qquad\text{if } e_{k+1}(i,j) = \mathbb{0}\text{ and }e_{k}(i,j) = \mathbb{0}\\
    \tau_k(\mathcal{G}_m), \\\qquad\qquad\text{otherwise}
\end{cases}
\label{eq:delay}
\end{multline}
where $e(i,j)$$=$$\mathbb{0}$ is weakly-connected edge, $e(i,j)$$=$$\mathbb{1}$ is strongly-connected edge; $ \tau_k(\mathcal{G}_m)$ is the cycle time at the $k$-th computation round during the training process. 
In general, using Eq.~\ref{eq:delay}, \textit{the delay of the next communication round $d_{k+1}$ is updated based on the delay of the previous rounds} and other factors, depending on the edge type connection.


\textbf{Cycle Time}. The cycle time per round is the time required to complete a communication round~\cite{marfoq2020throughput}. In this work, we define the cycle time per round is the maximum delay between all silo pairs with strongly-connected edges. Therefore, the average cycle time of the entire training is:
\begin{equation}
    \tau(\mathcal{G}_m) =\\ \frac{1}{k }\sum^{k-1}_{k=0} \left(\underset{j \in \mathcal{N}^{++}_{i} \cup\{i\}, \forall i \in \mathcal{V}}{\text{max}} \left(d_k\left(j,i\right)\right)\right),
\label{eq:shortly_cycle_time_ori}
\end{equation}
where $\mathcal{N}_{i}^{++}$ is an in-neighbors silo set of $i$ whose edges are strongly-connected.


\begin{figure*}[!ht]
  \centering
  \subfigure{\includegraphics[width=1.0\linewidth]{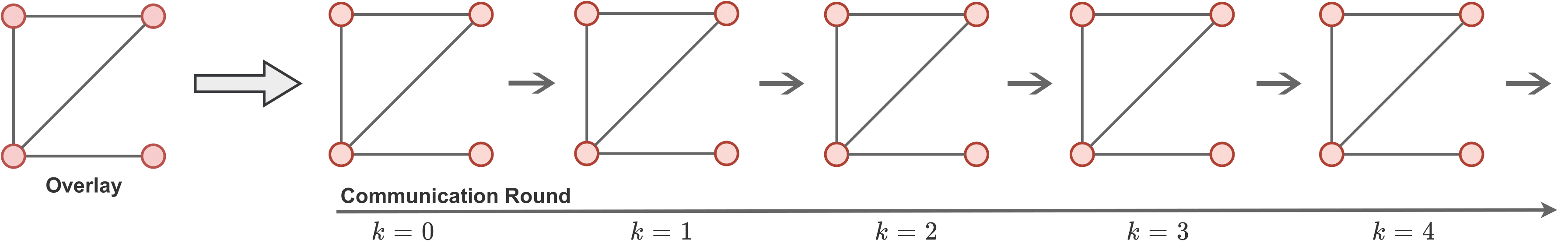}}
  \subfigure
   {\includegraphics[width=1.0\linewidth]{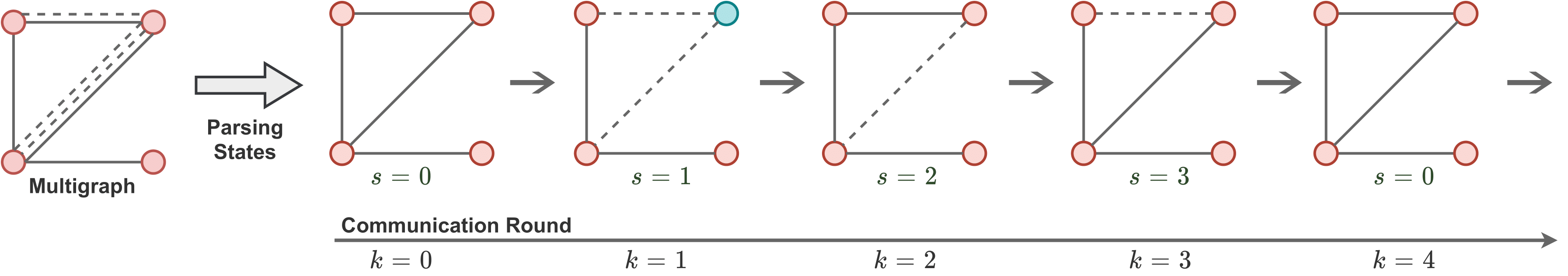}}
 \caption{ 
The comparison between RING~\cite{marfoq2020throughput} topology and our multigraph topology in each communication round. (a) RING uses the same overlay in each round. (b) Our proposed multigraph is parsed into different graph states. Each graph state is used in a communication round. Lines denote strongly-connected edges, dotted lines denote weakly-connected ones, and the blue color indicates isolated nodes.}
 \label{fig:learningProcess}
\vspace{-0.5cm}
\end{figure*}

\section{Multigraph Topology}
Our method first constructs the multigraph based on an overlay. Then we parse this multigraph into multiple states that may have isolated nodes. Note that, we do not choose isolated nodes randomly, but rely on the delay time. In practice, we observe that \textit{node with long delay time is the main reason for increasing the training time} since other nodes have to wait for it. Therefore, we want long delay nodes to become isolated nodes, and skip the model aggregation step on them. In the next communication round, the delay time of the isolated node will be updated using Eq.~\ref{eq:delay}, therefore it can become a normal node. This strategy allows us to reduce the waiting time in isolated nodes while ensuring that isolated nodes can become normal nodes and contribute to the training after a number of communication rounds.



\subsection{Multigraph Construction}
\SetKwInput{KwInput}{Input}                
\SetKwInput{KwOutput}{Output}              
\setlength{\algomargin}{1.0em}
\begin{algorithm}[!h]

\DontPrintSemicolon
\small
  \KwInput{Overlay $\mathcal{G}_o = (\mathcal{V}, \mathcal{E}_o)$; 
  \\\qquad \quad  
  Maximum edge between two nodes $t$.} 
  \KwOutput{Multigraph $\mathcal{G}_m = (\mathcal{V}, \mathcal{E}_m)$;
  \\\qquad \quad List number of edges between silo pairs $\mathcal{L}$.}
  
     \tcp{Delay computation for overlay}
     $D_o \leftarrow$ Establish a list of all delays to each silo pair.
     
     \ForEach{edge $e(i,j) \in \mathcal{E}_o$}{ 
     $d(i,j) \leftarrow$ Compute delay in overlay using Eq.~\ref{eq:ori_delay}.
      
     Append the computed delay $d(i,j)$ into $D_o$.
     }
     
     \tcp{Multigraph Establishment }
     
     $d_{\rm min} \leftarrow$ Compute the smallest delay by $\text{min}(D_0)$.
     
     $\mathcal{E}_m \leftarrow$ Establish the multiset containing all edges.
     
     $\mathcal{L}[|\mathcal{V}|,|\mathcal{V}|] \leftarrow$ Initialize the full-zero list for tracking the number of edges between each silo pair.
     
     \ForEach{edge $e(i,j) \in \mathcal{E}_o$}{ 
     $n(i,j) \leftarrow$ Find the number of edges for (i,j) pair by computing $\text{min}\left(t, \text{round}\left(\frac{d(i,j)}{d_{\text{min}}}\right)\right) $ 
     
     $\mathcal{E}_t \leftarrow$ Establish the set for each edge $e(i,j)$ that has marked with connection status. $\mathbb{1}$ labeled as strong-connected edge while $\mathbb{0}$ represented for weak-connected edge.
     
     Append  strong-connected edge $e(i,j)=\mathbb{1}$ into $\mathcal{E}_t$.
     
     \ForEach {$(n(i,j)-1)$}{Append  weak-connected edge $e(i,j)=\mathbb{0}$ into $\mathcal{E}_t$.}
     Append established edge set $\mathcal{E}_t$ into multiset $\mathcal{E}_m$.
    
     Update the number of edges between $(i,j)$ pair $n(i,j)$ into the corresponding position $\mathcal{L}[i,j]$ of track list $\mathcal{L}$.
     }
     
     \textbf{return} Multigraph $\mathcal{G}_m = (\mathcal{V}, \mathcal{E}_m)$; Track list $\mathcal{L}$

\caption{Multigraph Construction.}
\label{alg:strongly_weakly_edges}
\end{algorithm}

Algorithm~\ref{alg:strongly_weakly_edges} describes our methods to generate the multigraph $\mathcal{G}_m$ with multiple edges between silos. The algorithm takes the overlay $\mathcal{G}_o$ as input. Similar to~\cite{marfoq2020throughput}, we use the Christofides algorithm~\cite{monnot2003approximation} to obtain the overlay. In Algorithm~\ref{alg:strongly_weakly_edges}, we establish multiple edges that indicate different statuses (strongly-connected or weakly-connected). To identify the total edges between a silo pair, we divide the delay $d(i,j)$ by the smallest delay $d_{\min}$ overall silo pairs, and compare it with the maximum number of edges parameter $t$ ($t=5$ in our experiments). \textit{We assume that the silo pairs with longer delay will have more weakly-connected edges, hence potentially becoming isolated nodes}. Overall, we aim to increase the number of weakly-connected edges, which generate more isolated nodes to speed up the training process. Note that, from Algorithm~\ref{alg:strongly_weakly_edges}, each silo pair in the multigraph should have one strongly-connected edge and multiple weakly-connected edges. The role of the strongly-connected edge is to make sure that two silos have a good connection in at least one communication round.


\subsection{Multigraph Parsing}
\begin{algorithm}
\SetKwInput{KwInput}{Input}                
\SetKwInput{KwOutput}{Output}              
\DontPrintSemicolon
\small
  \KwInput{Multigraph $\mathcal{G}_m = (\mathcal{V}, \mathcal{E}_m)$;
  \\\qquad \quad List edge numbers between silo pairs $\mathcal{L}$.}
  \KwOutput{List of multigraph states $\mathcal{S} = \{\mathcal{G}_m^s = (\mathcal{V}, \mathcal{E}_m^s)\}$.}
     
     
     $s_{\rm max} \leftarrow$ Compute maximum number of distinct state in $\mathcal{G}_m$ by using Least Common Multiple~\cite{hardy1979introduction}

     
     $\bar{\mathcal{L}}\leftarrow$ Establish the dynamic list that tracks the number of edges between silo pairs during changing graph state and is initialized by input list edge $\mathcal{L}$.
     
     $\bar{\mathcal{E}}_m^s \leftarrow$ Establish the list of all possible extracted states.
     
     \tcp{States of Multigraph Establishment}

     \For{state $s = 0$ \KwTo $s_{\rm max}$}{
     
        $\mathcal{E}_t\leftarrow $ Establish temporary edge set.
        
        \ForEach{edge $e(i,j) \in \mathcal{E}_m$}{
        \If {$\bar{\mathcal{L}}[i,j] = \mathcal{L}[i,j]$}{Append strong connected-edge $e(i,j)=\mathbb{1}$ into edge set $\mathcal{E}_t$.}
        \Else {Append weak-connected edge $e(i,j)=\mathbb{0}$ into edge set $\mathcal{E}_t$.}
           
        \If {$\bar{\mathcal{L}}[i,j] = 1$}{Update the status of dynamic list $\bar{\mathcal{L}}$ using input $\mathcal{L}$ through $\bar{\mathcal{L}}[i,j] = \mathcal{L}[i,j]$.}
        \Else{Reduce the corresponding number of edges by $\bar{\mathcal{L}}[i,j]-= 1$.}
        }
        Append  edge set $\mathcal{E}_t$ into  $\bar{\mathcal{E}}_m^s$
     }
     
     \textbf{return} List of multigraph state $\mathcal{S} =\{\mathcal{G}_m^s = (\mathcal{V}, \mathcal{E}_m^s)\}$ by using list of possible extracted states $\bar{\mathcal{E}}_m^s$.
\caption{Multigraph Parsing.}
\label{alg:state_form}

\end{algorithm}

In Algorithm~\ref{alg:state_form}, we parse multigraph $\mathcal{G}_m$ into multiple graph states  $\mathcal{G}_m^s$. Graph states are essential to identify the connection status of silos in a specific communication round to perform model aggregation. In each graph state, our goal is to identify the isolated nodes. During the training, isolated nodes update their weights internally and ignore all weakly-connected edges that connect to them.

To parse the multigraph into graph states, we first identify the maximum of states in a multigraph $s_{\max}$ by using the least common multiple (LCM)~\cite{hardy1979introduction}. We then parse the multigraph into $s_{\max}$ states. The first state is always the overlay since we want to make sure all silos have a reliable topology at the beginning to ease the training. The reminding states are parsed so there is only one connection between two nodes. Using our algorithm, some states will contain isolated nodes. During the training process, only one graph state is used in a communication round.  Figure~\ref{fig:learningProcess} illustrates the training process in each communication round using multiple graph states.

\begin{table*}[t]
\centering
\setlength{\tabcolsep}{0.35 em} 
{\renewcommand{\arraystretch}{1.2}
\resizebox{0.9\textwidth}{!}{
\begin{tabular}{c|c|r|r|r|r|r|r|>{\columncolor[HTML]{EFEFEF}}c }
\hline
\multirow{2}{*}{\textbf{\begin{tabular}[c]{@{}c@{}}Dataset\end{tabular}}} &\multirow{2}{*}{\textbf{\begin{tabular}[c]{@{}c@{}}Network\end{tabular}}} & \multicolumn{7}{c}{\textbf{Topology Design}}\\ \cline{3-9} 
&  &\multicolumn{1}{|c|}{ \textbf{\small{STAR}}~\cite{brandes2008variants}} & \multicolumn{1}{|c|}{\textbf{\small{MATCHA}}~\cite{wang2019matcha}} & \multicolumn{1}{|c|}{\textbf{\small{MATCHA(+)}}~\cite{marfoq2020throughput}} & \multicolumn{1}{|c|}{\textbf{\small{MST}}~\cite{prim1957shortest}} & \multicolumn{1}{|c|}{\textbf{\small{$\updelta$-MBST}}~\cite{marfoq2020throughput}} & \multicolumn{1}{|c|}{\textbf{\small{RING}}~\cite{marfoq2020throughput}} &\textbf{\small{Ours}} \\ \hline
\multirow{5}{*}{\begin{tabular}[c]{@{}c@{}}\rotatebox[origin=c]{90}{\textbf{FEMNIST}}\end{tabular}} &Gaia                                                                                               & 289.8 \improvecolor($\downarrow$ 18.5)         & 166.4 \improvecolor($\downarrow$ 10.6)           & 166.4 \improvecolor($\downarrow$ 10.6)              & 77.2 \improvecolor($\downarrow$ 4.9)          & 77.2 \improvecolor($\downarrow$ 4.9)                   & 57.2 \improvecolor($\downarrow$ 3.6)          &  \textbf{15.7}         \\ 
&Amazon                                                                                            & 98.8 \improvecolor($\downarrow$ 7.3)          & 57.7 \improvecolor ($\downarrow 4.2$)            & 57.7 \improvecolor($\downarrow$ 4.2)               & 28.7 \improvecolor($\downarrow$ 2.1)        & 28.7 \improvecolor($\downarrow$ 2.1)                   & 20.3 \improvecolor($\downarrow$ 1.5)          & \textbf{13.6}          \\ 
&Géant                                                                                               & 132.2 \improvecolor($\downarrow$ 11.0)         & 46.9 \improvecolor($\downarrow$ 3.9)            & 102.3 \improvecolor($\downarrow$ 8.5)             & 40.1 \improvecolor($\downarrow$ 3.3)         & 40.1 \improvecolor($\downarrow$ 3.3)                    & 27.7 \improvecolor($\downarrow$ 2.3)          & \textbf{12.0}            \\ 
&Exodus                                                                                            & 265.2 \improvecolor($\downarrow$ 21.9)         & 84.7 \improvecolor($\downarrow$ 7.0)            & 211.5 \improvecolor($\downarrow$ 17.5)              & 84.4 \improvecolor($\downarrow$ 7.0)         & 84.4 \improvecolor($\downarrow$ 7.0)                    & 24.7 \improvecolor($\downarrow$ 2.0)          & \textbf{12.1}          \\ 
&Ebone                                                                                             & 190.9 \improvecolor($\downarrow$ 15.0)         & 61.5 \improvecolor($\downarrow$ 4.8)            & 112.6 \improvecolor($\downarrow$ 8.9)              & 60.9 \improvecolor($\downarrow$ 4.8)         & 60.9 \improvecolor($\downarrow$ 4.8)                    & 18.5 \improvecolor($\downarrow$ 1.5)          & \textbf{12.7}          \\ \hline \hline
\multirow{5}{*}{\begin{tabular}[c]{@{}c@{}}\rotatebox[origin=c]{90}{\textbf{iNaturalist}}\end{tabular}} &Gaia                                                                                                  & 390.9 \improvecolor($\downarrow$ 5.7)        & 227.4 \improvecolor($\downarrow$ 3.3)           & 227.4 \improvecolor($\downarrow$ 3.3)              & 138.1 \improvecolor($\downarrow$ 2.0)        & 138.1 \improvecolor($\downarrow$ 2.0)                   & 118.1 \improvecolor($\downarrow$ 1.7)         & \textbf{68.6}          \\ 
&Amazon                                                                                             & 288.1 \improvecolor($\downarrow$ 3.5)         & 123.9 \improvecolor($\downarrow$ 1.5)           & 123.9 \improvecolor($\downarrow$ 1.5)              & 89.7 \improvecolor($\downarrow$ 1.1)         & 89.7 \improvecolor($\downarrow$ 1.1)                    & \textbf{81.3} \improvecolor($\downarrow$ 1.0)          & \textbf{81.3}          \\
&Géant                                                                                                 & 622.3 \improvecolor($\downarrow$ 9.1)         & 107.9 \improvecolor($\downarrow$ 1.6)           & 452.5 \improvecolor($\downarrow$ 6.6)              & 101 \improvecolor($\downarrow$ 1.5)          & 101 \improvecolor($\downarrow$ 1.5)                     & 109.0 \improvecolor$\downarrow$ 1.6)           & \textbf{68.1}          \\ 
&Exodus                                                                                              & 911.9 \improvecolor($\downarrow$ 14.6)         & 145.7 \improvecolor($\downarrow$ 2.3)          & 593.2 \improvecolor($\downarrow$ 9.5)             & 145.3 \improvecolor($\downarrow$ 2.3)       & 145.3 \improvecolor($\downarrow$ 2.3)                  & 103.9 \improvecolor($\downarrow$ 1.7)        & \textbf{62.6}          \\
&Ebone                                                                                              & 901.7 \improvecolor($\downarrow$ 13.9)        & 122.5 \improvecolor($\downarrow$ 1.9)          & 579.9 \improvecolor($\downarrow$ 8.9)             & 121.8 \improvecolor($\downarrow$ 1.9)       & 121.8 \improvecolor($\downarrow$ 1.9)                  & 95.3 \improvecolor($\downarrow$ 1.5)         & \textbf{64.9}          \\ \hline \hline
\multirow{5}{*}{\begin{tabular}[c]{@{}c@{}}\rotatebox[origin=c]{90}{\textbf{Sentiment140}}\end{tabular}} &Gaia                                                                                                & 323.8 \improvecolor($\downarrow$ 10.5)         & 186.0 \improvecolor($\downarrow$ 6.0)            & 186 \improvecolor($\downarrow$ 6.0)               & 96.8 \improvecolor($\downarrow$ 3.1)        & 96.8 \improvecolor($\downarrow$ 3.1)                   & 76.8 \improvecolor($\downarrow$ 2.5)         & \textbf{31.0}          \\ 
&Amazon                                                                                        & 164.6 \improvecolor($\downarrow$ 4.6)        & 79.2 \improvecolor($\downarrow$ 2.2)           & 79.2 \improvecolor($\downarrow$ 2.2)              & 48.4 \improvecolor($\downarrow$ 1.4)         & 48.4 \improvecolor($\downarrow$ 1.4)                   & 40.0 \improvecolor($\downarrow$ 1.1)         & \textbf{35.8}          \\ 
&Géant                                                                                               & 310.5 \improvecolor($\downarrow$ 10.3)        & 66.6 \improvecolor($\downarrow$ 2.2)           & 222.6 \improvecolor($\downarrow$ 7.4)             & 59.7 \improvecolor($\downarrow$ 2.0)        & 59.7 \improvecolor($\downarrow$ 2.0)                    & 54.9 \improvecolor($\downarrow$ 1.8)          & \textbf{30.3}          \\
&Exodus                                                                                             & 495.4 \improvecolor($\downarrow$ 17.7)        & 104.3 \improvecolor($\downarrow$ 3.7)          & 346.3 \improvecolor($\downarrow$ 12.4)             & 104.1 \improvecolor($\downarrow$ 3.7)       & 104.1 \improvecolor($\downarrow$ 3.7)                  & 50.6 \improvecolor($\downarrow$ 1.8)         & \textbf{28.0}          \\ 
&Ebone                                                                                              & 444.2 \improvecolor($\downarrow$ 15.3)        & 81.1 \improvecolor($\downarrow$ 2.8)           & 262.2 \improvecolor($\downarrow$ 9.0)             & 80.5 \improvecolor($\downarrow$ 2.8)        & 80.5 \improvecolor($\downarrow$ 2.8)                   & 43.9 \improvecolor($\downarrow$ 1.5)         & \textbf{29.1}          \\ \hline
\end{tabular}
}
\vspace{1ex}
\caption{Cycle time (ms) comparison between different typologies. ($\improvecolor \downarrow$ $\circ$) indicates our reduced times compared with other methods.
\label{tab:cycle_time_FULL}
}
}
\end{table*}

\subsection{Multigraph Training} In each communication round, a state graph $\mathcal{G}_m^s$ is selected in a sequence that identifies the topology design used for training. We then collect all strongly-connected edges in the graph state $\mathcal{G}_m^s$ in such a way that nodes with strongly-connected edges need to wait for neighbors, while the isolated ones can update their models. We train our multigraph with DPASGD algorithm~\cite{wang2018cooperative}:
\begin{equation}
\small
\textbf{w}_{i}\left(k + 1\right) =
\begin{cases}
    \sum_{j \in \mathcal{N}_{i}^{++} \cup{\{i\}}}\textbf{A}_{i,j}\textbf{w}_{j}\left(k - h\right), \\\qquad\qquad\quad \text{if k} \equiv 0 \left(\text{mod }u + 1\right) \text{\&} \left|\mathcal{N}_{i}^{++}\right| > 1 ,\\
    \textbf{w}_{i}\left(k\right)-\alpha_{k}\frac{1}{b}\sum^b_{h=1}\nabla L_i\left(\textbf{w}_{i}\left(k\right),\xi_i^{\left(h\right)}\left(k\right)\right), \\\qquad\qquad\quad\text{otherwise.}
\end{cases}
\label{eq:skipbatch_DFL}
\end{equation}
where $(k- h)$ is the index of the considered weights; 
$h$ is initialized to $0$ and 
$h = h + 1, \: \text{if } \:\: e_{k-h}(i,j) = \mathbb{0}$. Through Eq.~\ref{eq:skipbatch_DFL}, at each state, if a silo is not an isolated node, it must wait for the model from its neighbor to update its weight. If a silo is an isolated node, it can use the model in its neighbor from the $(k-h)$ round to update its weight immediately.

\section{Experiments}
\subsection{Experimental Setup}
\textbf{Datasets.} We use three datasets in our experiments: Sentiment140~\cite{go2009twitter}, iNaturalist~\cite{van2018inaturalist}, and FEMNIST~\cite{caldas2018leaf}. All datasets and the pre-processing process are conducted by following recent works~\cite{wang2019matcha} and~\cite{marfoq2020throughput}. Table~\ref{tab:Dataset_Analysis1} shows the dataset setups in detail.

\begin{table}[ht]
\centering
\setlength{\tabcolsep}{0.35 em} 
\renewcommand{\arraystretch}{1.2}
\caption{Dataset statistic and model details in our experiments. The model size is in Mbits.}
\vspace{2ex}
\label{tab:Dataset_Analysis1}
\begin{tabular}{c|c|c|c}
\hline
\textbf{Dataset}                            & \textbf{\small{FEMNIST}} & \textbf{\small{Sentiment140}}                                             & \textbf{\small{iNaturalist}} \\ \hline
\#Samples                            & 805M                                                         & 1,600M                                                                                                       & 450M                                                                 \\ \hline
Model                              & CNN~\cite{marfoq2020throughput}                                                 &  LSTM~\cite{hochreiter1997long} & ResNet~\cite{he2016deep}                 \\ \hline

\#Params                           & 1,2M                                                         & 4,8M                                                                                                         & 11,2M                                                                \\ \hline

Batch size                         & 128                                                          & 512                                                                                                          & 16                                                                   \\ \hline
Model size                         & 4.62                                                         & 18.38                                                                                                        & 42.88                                                                \\ \hline
\end{tabular}

\end{table}

\textbf{Network}.
Following~\cite{marfoq2020throughput}, we consider five distributed networks in our experiments: Exodus, Ebone, Géant, Amazon~\cite{awscloud} and Gaia~\cite{hsieh2017gaia}. The Exodus, Ebone, and Géant are from the Internet Topology Zoo~\cite{knight2011internetzoo}. The Amazon and Gaia network are synthetic and are constructed using the geographical locations of the data centers. 

\textbf{Baselines}.
We compare our multigraph topology with recent state-of-the-art topology designs for federated learning: STAR~\cite{brandes2008variants}, MATCHA~\cite{wang2019matcha}, MATCHA(+)~\cite{marfoq2020throughput}, MST~\cite{prim1957shortest}, $\updelta$-MBST~\cite{marfoq2020throughput}, and RING~\cite{marfoq2020throughput}. 

\textbf{Hardware Setup}. Since measuring the cycle time is crucial to compare the effectiveness of all topologies in practice, we test and report the cycle time of all baselines and our method on the same NVIDIA Tesla P100 16Gb GPUs. No overclocking is used. 

\textbf{Time Simulator}. Similar to~\cite{marfoq2020throughput}, we adapted PyTorch with the MPI backend. We take advantage of the network simulator and the timing simulator as in Marfod \etal~\cite{marfoq2020throughput}. More details of our experimental setup and time simulator for computing wall-clock time can be found in our Supplementary Material.


\subsection{Cycle Time Comparison}

Table \ref{tab:cycle_time_FULL} shows the cycle time of our method in comparison with other recent approaches. This table illustrates that our proposed method significantly reduces the cycle time in all setups with different networks and datasets. In particular, compared to the state-of-the-art RING~\cite{marfoq2020throughput}, our method reduces the cycle time by $2.18$, $1.5$, $1.74$ times in average in the FEMNIST, iNaturalist, Sentiment140 dataset, respectively. Our method also clearly outperforms MACHA, MACHA(+), and MST by a large margin. The results confirm that our multigraph with isolated nodes helps reduce the cycle time in federated learning.

From Table~\ref{tab:cycle_time_FULL}, our multigraph achieves the minimum improvement under the Amazon network in all three datasets. This can be explained that, under the Amazon network, our proposed topology does not generate many isolated nodes. Hence, the improvement is limited. Intuitively, when there are no isolated nodes, our multigraph will become the overlay, and the cycle time of our multigraph will be equal to the cycle time of the overlay in RING.

\subsection{Isolated Node Analysis}

\textbf{Isolated Nodes vs. Network Configuration}. The numbers of isolated nodes vary based on the network configuration (Amazon, Gaia, Exodus, etc.). The parameter $t$ (maximum number of edges between two nodes), and the delay time which is identified by many factors (geometry distance, model size, computational cost based on tasks, bandwidth, etc. - please Eq.~\ref{eq:ori_delay}) also affect the process of generating isolated nodes. Table~\ref{tab:isolatedNodesVSNetworkConfig} illustrates the effectiveness of isolated nodes in different network configurations. Specifically, we conduct experiments on the FEMNIST dataset using five network configurations (Gaia, Amazon, Geant, Exodus, Ebone). We can see that our cycle time compared with RING~\cite{marfoq2020throughput} is reduced significantly when more communication rounds or graph states have isolated nodes.

\begin{table}[t]
\centering
\setlength{\tabcolsep}{0.2 em} 
{\renewcommand{\arraystretch}{1.2}
\begin{tabular}{c|c|c|r|r}
\hline
{\textbf{Network}} & {\textbf{\begin{tabular}[c]{@{}c@{}}Total\\ silos\end{tabular}}} & {\textbf{\begin{tabular}[c]{@{}c@{}}\#Rounds\\ \end{tabular}}} & \multicolumn{1}{|c|}{{\textbf{\begin{tabular}[c]{@{}c@{}}\#States\\ \end{tabular}}}} & {\textbf{\begin{tabular}[c]{@{}c@{}}Cycle Time \\ (ms)\end{tabular}}} \\ \hline
Gaia & 11 & 4693/6400 & 44/60  \color[HTML]{6200C9}(73.3\%) & 15.7 \improvecolor($\downarrow$3.6) \\ \hline
Amazon & 22 & 2133/6400  & 2/6 \color[HTML]{6200C9}(33.3\%)& 13.6 \improvecolor($\downarrow$1.5) \\ \hline
Géant & 40 & 4266/6400 & 8/12 \color[HTML]{6200C9}(66.7\%) & 12.0 \improvecolor($\downarrow$2.3) \\ \hline
Exodus & 79 & 3306/6400  & 31/60 \color[HTML]{6200C9}(51.7\%)& 12.1 \improvecolor($\downarrow$2.0) \\ \hline
Ebone & 87 & 2346/6400   & 11/30 \color[HTML]{6200C9}(36.7\%) & 12.7 \improvecolor($\downarrow$1.5) \\ \hline
\end{tabular}
\caption{The effectiveness of isolated nodes under different network configurations. All experiments are trained with $6400$ communication rounds on FEMNIST dataset.
We then record the number of states and rounds that have the appearance of isolated nodes and compare our cycle time with RING~\cite{marfoq2020throughput}. 
\label{tab:isolatedNodesVSNetworkConfig}
}
}
\end{table}

\textbf{Isolated Nodes vs. RING vs. Random Strategy.}
Isolated nodes play an important role in our method as we can skip the model aggregation step in the isolated nodes. 
In practice, we can have a trivial solution to create isolated nodes by randomly removing some nodes from the overlay of RING. Table~\ref{tab:isolatedNodes} shows the experiment results in two scenarios on FEMNIST dataset and Exodus Network: \textit{i)} Randomly remove some silos in the overlay of RING, and \textit{ii)} Remove the most inefficient silos (i.e., silos with the longest delay) in the overlay of RING. 
From Table~\ref{tab:isolatedNodes}, the cycle time reduces significantly when two aforementioned scenarios are applied. However, the accuracy of the model also drops significantly. This experiment shows that although randomly removing some nodes from the overlay of RING is a trivial solution, it can not maintain model accuracy. On the other hand, our multigraph not only reduces the cycle time of the model but also preserves the accuracy. This is because our multigraph can skip the aggregation step of the isolated nodes in a communication round. However, in the next round, the delay time of these isolated nodes will be updated, and they can become normal nodes and contribute to the final model.

\textbf{Isolated Nodes Illustration.} 
Figure~\ref{fig:IsoNode} shows a detailed illustration of our algorithm with the isolated nodes in a real-world training scenario. The experiment is conducted on Gaia network~\cite{hsieh2017gaia} geometry and their corresponding hardware for supporting link latency computation. The image classification task is chosen for this benchmarking by using FEMNIST dataset~\cite{caldas2018leaf} and CNN backbone provided by Marfod \etal~\cite{marfoq2020throughput}. Hence, we keep the model transmitted size at $4.62$ Mb, all access links have $10$  Gbps traffic capacity, the number of local updates is set to $1$, and the maximum number of edges $t$ is set to $3$. As shown in Figure~\ref{fig:IsoNode}, although there are no isolated nodes in the initialized state, the number of isolated nodes increases in the next consequence states with a vast number (4 nodes). This circumstance leads to a $\sim 4$ times reduction in cycle time compared to the initialized state. The appearance of isolated nodes also greatly reduces the connection between silos by $\sim 3.6$ times, from $11$ down to $3$
 connections, and also discarded ones all have high latency.

\begin{table}[t]
\centering
\setlength{\tabcolsep}{0.15 em} 
{\renewcommand{\arraystretch}{1.1}
\begin{tabular}{c|c|c|c|c}
\hline
\textbf{Methods}                & \textbf{Criteria}                                                                                      & \multicolumn{1}{c|}{\textbf{\begin{tabular}[c]{@{}c@{}}\#Removed\\Nodes\end{tabular}}} & \textbf{\begin{tabular}[c]{@{}c@{}}Cycle\\ Time (ms)\end{tabular}} & \textbf{Acc (\%)} \\ \hline
\multirow{9}{*}{{RING~\cite{marfoq2020throughput}}} & \textit{Baseline}                                                                                      & \_                                                                                  & 24.7                                                          & 71.05             \\ \cline{2-5} 
                                & \multirow{4}{*}{\textit{\begin{tabular}[c]{@{}c@{}}Randomly \\ remove silos \\ in overlay\end{tabular}}} & 1                                                                                   & 23.1                                                   & 70.63     \\ \cline{3-5} 
                                &                                                                                                        & 5                                                                                   & 21.7                                                  & 68.57    \\ \cline{3-5} 
                                &                                                                                                        & 10                                                                                  & 18.8                                                   & 64.23     \\ \cline{3-5} 
                                &                                                                                                        & 20                                                                                  & 13.0                                                   & 61.2     \\ \cline{2-5} & \multirow{4}{*}{\textit{\begin{tabular}[c]{@{}c@{}}Remove most \\ inefficient\\ silos\end{tabular}}}   & 1                                                                                   & 22.5                                         & 70.71    \\ \cline{3-5} 
                                &                                                                                                        & 5                                                                                   & 19.5                                         & 68.37   \\ \cline{3-5} 
                                &                                                                                                        & 10                                                                                  & 15.8                                         & 63.13    \\ \cline{3-5} 
                                &                                                                                                        & 20                                                                                  & 11.2                                         & 61.48    \\
                                 \hline
\rowcolor[HTML]{EFEFEF} \begin{tabular}[c]{@{}c@{}}Multigraph \\(ours)\end{tabular}                   & \_                                                                                                     & \_                                                                                  & \textbf{12.1}                                                 & \textbf{71.13}    \\ 
\hline
\end{tabular}
\caption{The cycle time and accuracy of our multigraph vs. RING with different criteria. 
\label{tab:isolatedNodes}
}
}
\end{table}


\begin{figure}[t]
  \centering
\resizebox{\linewidth}{!}{
\setlength{\tabcolsep}{0.5pt}
\begin{tabular}{cc}
\shortstack{\includegraphics[width=0.45\linewidth,height=0.37\linewidth]{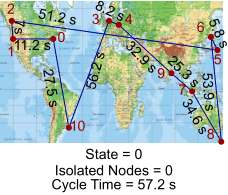}}&
\shortstack{\includegraphics[width=0.45\linewidth,height=0.37\linewidth]{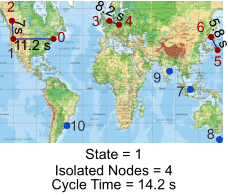}}\\
\shortstack{\includegraphics[width=0.45\linewidth,height=0.37\linewidth]{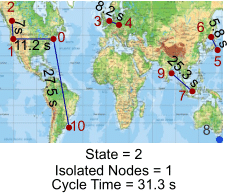}}&
\shortstack{\includegraphics[width=0.45\linewidth,height=0.37\linewidth]{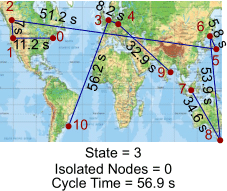}}\\
\shortstack{\includegraphics[width=0.45\linewidth,height=0.37\linewidth]{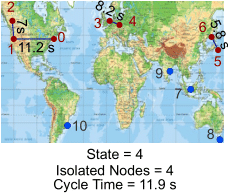}}&
\shortstack{\includegraphics[width=0.45\linewidth,height=0.37\linewidth]{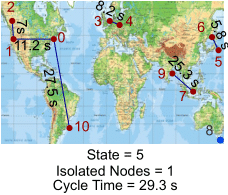}}\\
\end{tabular}
}
    \caption{Isolated nodes illustration. Red nodes indicate normal nodes. Blue nodes are isolated nodes.
    }
 \label{fig:IsoNode}
\end{figure}

\subsection{Multigraph Ablation Study}
\textbf{Accuracy Analysis.}
In federated learning, improving the model accuracy is not the main focus of topology designing methods. However, preserving the accuracy is also important to ensure model convergence. Table~\ref{tab:network_stability_analysis_FEMNIST} shows the accuracy of different topologies after $6,400$ communication training rounds on the FEMNIST dataset. This table illustrates that our proposed method achieves competitive accuracy with other topology designs. This confirms that our topology can maintain the accuracy of the model, while significantly reducing the training time.

\begin{figure}[]
  \centering
    \subfigure[Train Loss]{\includegraphics[width=0.44\linewidth]{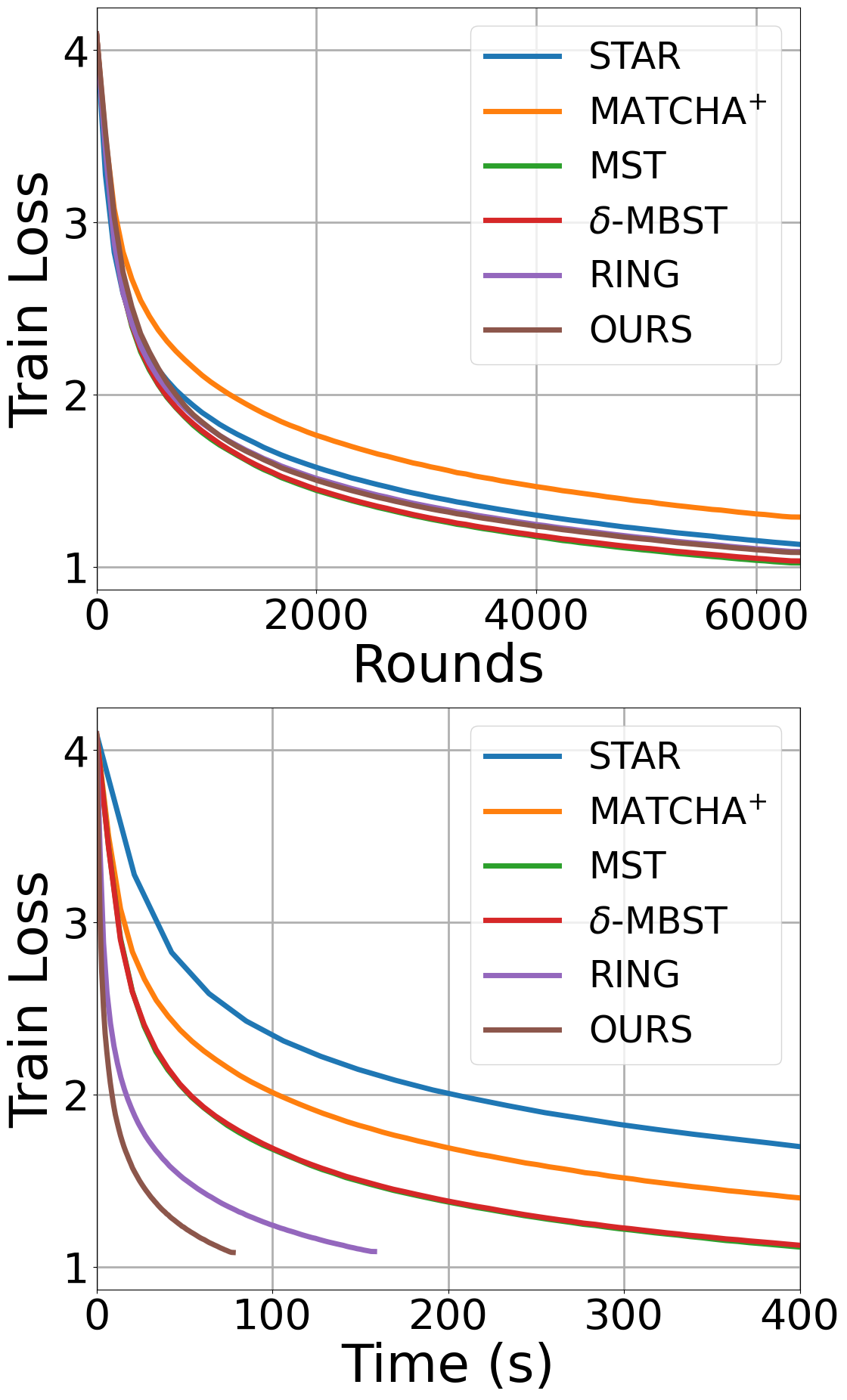}}\hspace{1ex}
    \subfigure[Train Accuracy]{\includegraphics[width=0.48\linewidth]{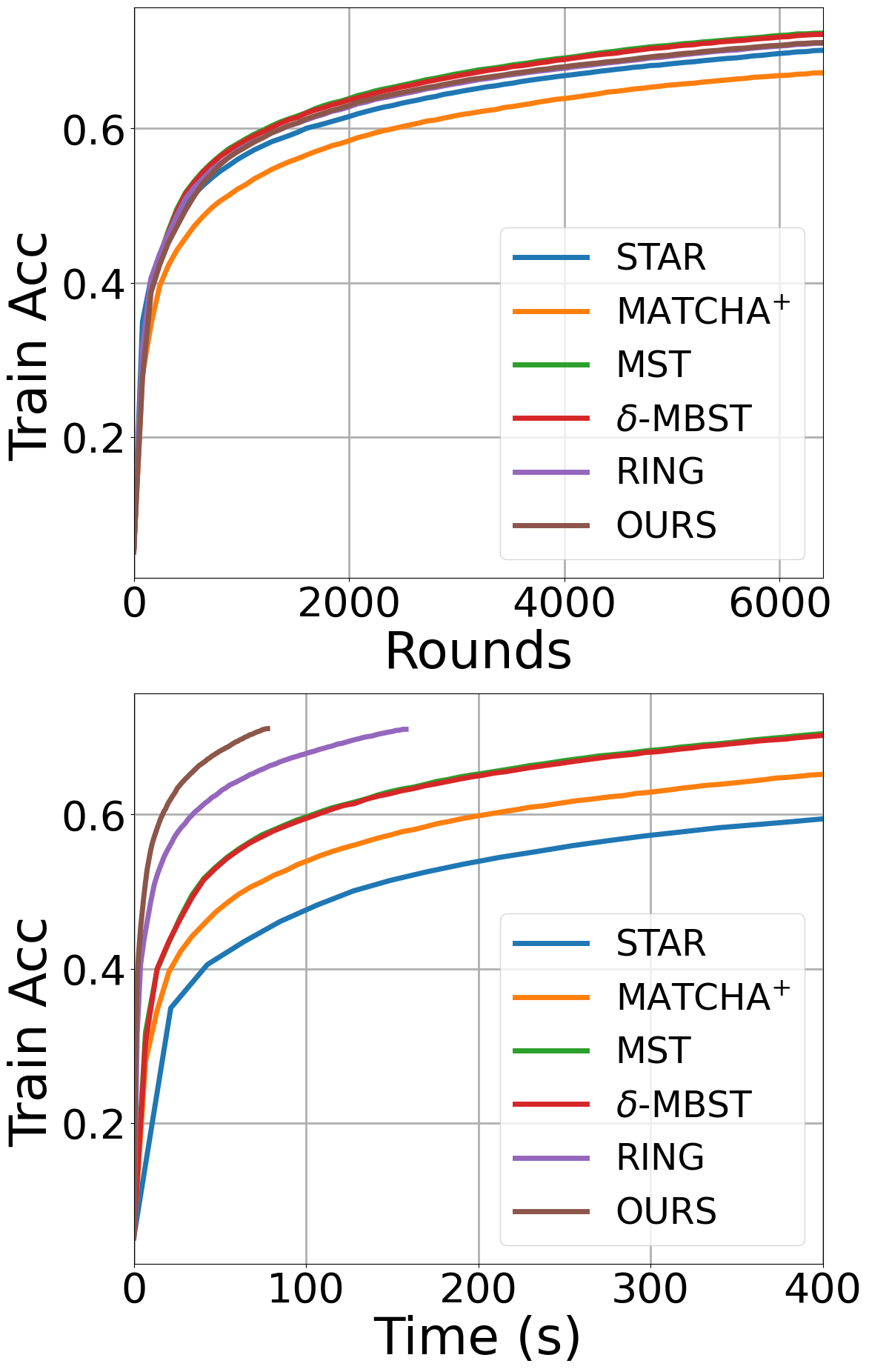}}
 \caption{Convergence analysis of our multigraph under communication rounds (top row) and wall-clock time (bottom row). All access links have a 10 Gbps capacity. The training time is counted after all setups finish $6,400$ communication rounds.}
 \label{fig:convergence}
\end{figure}

\textbf{Convergence Analysis}. Figure~\ref{fig:convergence} shows the training loss versus the number of communication rounds and the wall-clock time under Exodus network using the FEMNIST dataset. This figure illustrates that our proposed topology converges faster than other methods while maintaining the model accuracy. 
We observe the same results in other datasets and network setups.


\begin{table}[t]
\centering
\setlength{\tabcolsep}{0.25 em} 
{\renewcommand{\arraystretch}{1.2}
\resizebox{1.0\textwidth}{!}{\begin{tabular}{c|c|c|c|c|c|>{\columncolor[HTML]{EFEFEF}}c}
\hline
\multirow{2}{*}{\textbf{\begin{tabular}[c]{@{}c@{}}Network\end{tabular}}} & \multicolumn{6}{c}{\textbf{Topology}}     \\ \cline{2-7} 
     & \textbf{\small{STAR}} & \textbf{\small{MATCHA(+)}} & \textbf{\small{MST}} & \textbf{\small{$\updelta$-MBST}} & \textbf{\small{RING}} & \textbf{\small{Ours}} \\ \hline
Gaia                                                                    & 69.09         & 68.43              & 68.86        & 68.95           & 68.2          & 68.45        \\ 
Amazon                                                                  & 69.59         & 69.06              & 69.65        & 70.37           & 69.78         & 69.63        \\ 
Géant                                                                   & 68.91         & 65.57              & 69.44        & 68.94           & 69.3          & 68.98        \\
Ebone                                                                  & 69.66         & 64.48              & 71.91        & 70.62           & 70.29         & 70.23        \\ 
Exodus                                                                   & 70.14         & 67.21              & 72.36        & 72.19           & 71.05         & 71.13        \\ \hline
\end{tabular}
}
\caption{Accuracy comparison between different topologies. The experiment is conducted using the FEMNIST dataset. The accuracy is reported after $6,400$ communication rounds in all methods.
\label{tab:network_stability_analysis_FEMNIST}
}
}
\end{table}


\begin{table}[t]
\centering
\setlength{\tabcolsep}{0.4 em} 
{\renewcommand{\arraystretch}{1.1}
\vspace{2ex}

\begin{tabular}{c|c|c|c}
\hline
\textbf{Topology}             & \textbf{$t$} & \textbf{Cycle time  (ms)} & \textbf{Acc(\%)} \\
\hline
\multirow{1}{*}{RING~\cite{marfoq2020throughput}} & \_             & 24.7             & 71.05   \\\hline
\multirow{7}{*}{{\begin{tabular}[c]{@{}c@{}}Multigraph\\ (ours)\end{tabular}}} & 1             & 24.7             & 71.05   \\ \cline{2-4} 
                                     & 3             & 13.5                      & 71.08            \\ \cline{2-4} 
                                     & 5             & 12.1                      & 71.13            \\ \cline{2-4} 
                                     & 8             & 11.9                      & 69.27            \\ \cline{2-4} 
                                     & 10            & 11.9                      & 69.27            \\ \cline{2-4} 
                                     & 20            & 11.9                      & 69.27 
                                     \\ \cline{2-4} 
                                     & 30            & 11.9                      & 69.27  \\ \hline
\end{tabular}
\caption{Cycle time and accuracy trade-off with different value of $t$, i.e., the maximum number of edges between two nodes.
\label{tab:t_max_analysis}
}
}
\end{table}

\textbf{Cycle Time and Accuracy Trade-off.} In our method, the maximum number of edges between two nodes $t$ in Algorithm~\ref{alg:strongly_weakly_edges} mainly affects the number of isolated nodes. This leads to a trade-off between the model accuracy and cycle time. Table~\ref{tab:t_max_analysis} illustrates the effectiveness of this parameter. When $t = 1$, we technically consider there are no weak connections and isolated nodes. Therefore, our method uses the original overlay from RING. When $t$ is set higher, we can increase the number of isolated nodes, hence decreasing the cycle time. In practice, too many isolated nodes will limit the model weights to be exchanged between silos. Therefore, models at isolated nodes are biased to their local data and consequently affect the final accuracy. 

\section{Conclusion}
We proposed a new multigraph topology for cross-silo federated learning. Our method first constructs the multigraph using the overlay. Different graph states are then parsed from the multigraph and used in each communication round. Our method significantly reduces the cycle time by allowing the isolated nodes in the multigraph to do model aggregation without waiting for other nodes. The intensive experiments on three datasets show that our proposed topology achieves new state-of-the-art results in all network and dataset setups.
\label{Sec:Conclusion}

{\small
\bibliographystyle{ieee_fullname}
\bibliography{egbib}
}

\end{document}

%% file: math_commands.tex
\usepackage{amsmath,amsfonts,bm,cuted}
\usepackage[ruled,vlined,linesnumbered]{algorithm2e}
\usepackage{graphicx}
\usepackage{tabularx} 
\usepackage{amssymb}
\usepackage{multirow}
\usepackage{floatrow}
\usepackage{subfigure}
\usepackage{dsfont}
\usepackage{upgreek}
\usepackage{mathrsfs}
\usepackage{url}
\usepackage{color}

\usepackage{pifont}
\def\eqref#1{equation~\ref{#1}}

\def\1{\bm{1}}

\DeclareMathAlphabet{\mathsfit}{\encodingdefault}{\sfdefault}{m}{sl}
\SetMathAlphabet{\mathsfit}{bold}{\encodingdefault}{\sfdefault}{bx}{n}

